# SACA: A Scenario-Aware Collision Avoidance Framework for Autonomous Vehicles Integrating LLMs-Driven Reasoning

Shiyue Zhao, Junzhi Zhang, Neda Masoud, Heye Huang, Xiaohui Hou, and Chengkun He

***Abstract*—Reliable collision avoidance under extreme situations remains a critical challenge for autonomous vehicles. While large language models (LLMs) offer promising reasoning capabilities, their application in safety-critical evasive maneuvers is limited by latency and robustness issues. Even so, LLMs stand out for their ability to weigh emotional, legal, and ethical factors, to support socially responsible and context-aware collision avoidance decision-making. This paper proposes a scenario-aware collision avoidance (SACA) framework for extreme situations that integrates predictive scenario evaluation, data-driven reasoning, and scenario-preview-based deployment to improve collision avoidance decision-making. SACA consists of three key components. First, a predictive scenario analysis module utilizes obstacle reachability analysis and motion intention prediction to construct a comprehensive situational prompt. Second, an online reasoning module refines decision-making by leveraging prior collision avoidance knowledge and fine-tuning with scenario data. Third, an offline evaluation module assesses performance and stores scenarios in a memory bank. Additionally, A precomputed policy method improves deployability by previewing scenarios and retrieving or reasoning policies based on similarity and confidence levels. Real-vehicle tests show that, compared with baseline methods, SACA effectively reduces collision losses in extreme high-risk scenarios and lowers false triggering under complex conditions. Project page: https://sean-shiyuez.github.io/SACA/.**

## I. INTRODUCTION

Road safety remains a pressing global issue, with traffic accidents casting a long shadow over individuals, communities, and economies alike [1,2]. Autonomous driving systems, armed with sharp perception and proactive safety measures, have sparked hope for reducing these risks. Studies show they can handle routine driving with impressive precision, yet one stubborn challenge lingers: extreme collision-risk scenarios [3].

Imagine an **extreme scenario**: a self-driving car speeding toward an intersection as the light turns green. Suddenly, a truck barrels through a fading yellow from the left, while pedestrians inch across the crosswalk ahead (see Figure 1). What's the right move? Traditional collision avoidance systems (CAS)—whether rule-based [4] or powered by reinforcement learning (RL) [5]—often falter here, tethered to rigid mathematical frameworks that can't quite keep up.

These systems hinge on clear-cut objectives: rule-based approaches lean on preset thresholds to trigger actions [6], while RL methods encode priorities in reward functions [7]. Fine for predictable roads, but in chaotic, high-stakes moments, these setups hit a wall. Mathematical models alone struggle to juggle the tangled mix of physical dynamics, legal boundaries, and ethical dilemmas—like whether to swerve and risk the pedestrians or brake and face the truck. Even end-to-end autonomous driving systems, built on deep neural networks that map camera feeds straight to steering, stumble in these unseen scenarios [8]. They shine in everyday traffic but flinch at rare, messy situations—think icy roads or high-speed near-misses—where training data is scarce [9]. Plus, their black-box nature makes it nearly impossible to weave in broader concerns, like societal norms or legal accountability, when lives hang in the balance [10].

So, here's the question: **How can we develop a reliable collision avoidance system for extreme collision avoidance scenarios that integrates safety, legal, and ethical considerations, while remaining responsive during emergencies?** Rule-based logic can't bend for the unexpected, RL needs endless trial-and-error to learn rare events, and neural networks hide their reasoning in a fog. We need a smarter approach—one that merges the collision avoidance instincts of experts with a more general awareness of the wider real world.

This is where our approach turns to large language models (LLMs) for a fresh angle. Our idea is to tap into LLMs for scenario-aware reasoning, pairing them with a preview-based strategy to boost real-world deployability. Recent models like ChatGPT-4 and Grok have shown off their knack for grasping **general world knowledge, human emotions, and near-human reasoning skills.** Research backs this up—take GPT-4, for instance, which has demonstrated human-like judgment in ethical dilemmas, as seen in moral scenario evaluations, and even outpaced humans in empathy, according to a 2024 study [11]. What's more, LLMs stand out because they can spell out their thought process step-by-step, offering a window into their decisions that end-to-end neural networks can't match..

Still, two key challenges stand out:

**- How can we incorporate collision avoidance expertise and experience—like analyzing obstacles and predicting motion—into LLMs' reasoning?**

S. Zhao, J. Zhang, and C. He are with School of Vehicle and Mobility, Tsinghua University, Beijing 100084 China (+1-734-210-9574; e-mail: thu_znqzd@ 163.com).

S. Zhao is now invited as a visiting scholar in the Department of Civil and Environmental Engineering, University of Michigan, Ann Arbor, MI 48105 USA.

N. Masoud and X. Xia are with the Department of Civil and Environmental Engineering, University of Michigan, Ann Arbor, MI 48105 USA.

H. Huang is with the Connected & Autonomous Transportation Systems Lab, University of Wisconsin–Madison.

X. Hou is with the School of Automation at the Beijing Institute of Technology.

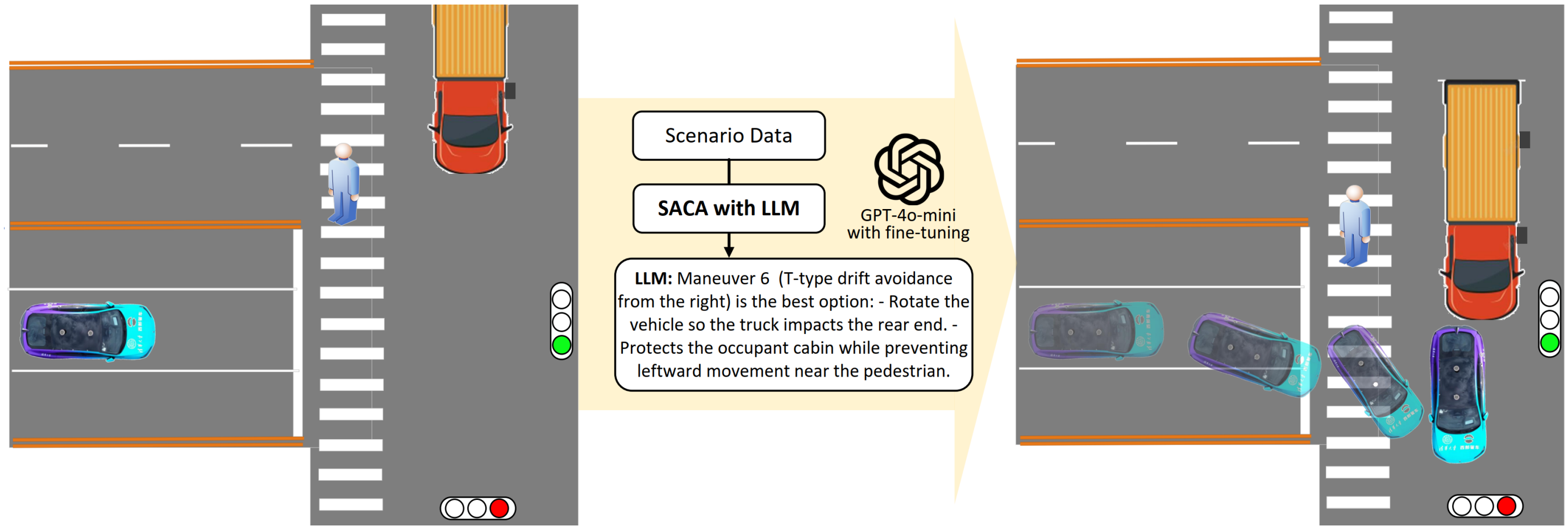

**Fig. 1**: High-risk intersection scenario with an autonomous vehicle (in blue) facing a truck (in red) and pedestrians. SACA, using GPT-4o-mini with fine-tuning, executes a T-type drift maneuver, shifting right to protect the cabin and avoid the pedestrian, as shown in the right panel.

**- How do we adjust these models to handle safety-critical scenarios where fast decisions are essential and mistakes can't happen?**

To address these, we introduce the Scenario-Aware Collision Avoidance (SACA) framework. Built on risk prediction, data-driven logic, and scenario previews, SACA seeks to enhance decisions and ensure real-world reliability

To the best of our knowledge, this work makes three key contributions:

**(1)** **Robust LLMs Adaptation via Fine-Tuning:** By fine-tuning with memory-based retrieval, our framework reliably handles safety-critical extreme scenarios and enables continuous improvements.

**(2)** **Scenario-Aware Risk Assessment for Collision Avoidance:** Through obstacle reachability analysis and traffic participant intention prediction, we construct a contextual model to guide LLMs.

**(3)** **Scenario-Preview-Based Deployment:** SACA leverages a memory bank of scenario and precomputed policies for low-latency deployment, adapting to extreme situations.

Figure 1 depicts a high-risk intersection scenario, as described earlier, where an autonomous vehicle (in blue) speeds toward a green light, facing an imminent threat from a truck (in red) running a yellow light from the left, with pedestrians crossing nearby. Thanks to SACA's LLM-driven reasoning—specifically GPT-4o-mini with fine-tuning—the vehicle executes a T-type drift avoidance maneuver, shifting right to let the truck impact the rear, thus protecting the occupant cabin and avoiding harm to the pedestrian, as shown in the right panel.

## II. RELATED WORK

**Risk Assessment and Motion Intention Recognition:** Research on obstacle and traffic participant risk assessment in autonomous driving relies on key theoretical frameworks, each with distinct challenges. The artificial potential field (APF) method, pioneered by Khatib [12], models risks using potential fields to represent obstacles as repulsive forces and goals as attractive ones, with Pan et al. [13] and Huang et al. [14] extending this to integrate road, vehicle, and velocity dynamics for collision avoidance [15]. Still, these approaches tend to overlook aspects like the ego vehicle's execution capability and its operational status.

Another approach, Hamilton–Jacobi (HJ) reachability analysis, offers a rigorous framework by identifying backward reachable sets—states where collisions with obstacles or traffic participants become unavoidable—while accounting for ego vehicle dynamics [16]. Li et al. [17] enhance HJ reachability by clustering vehicle behaviors (e.g., left/right turns) via trajectory prediction, yielding less conservative HJ-based controllers tested in T-intersections and roundabouts. Still, its heavy computational demands in high-dimensional settings, known as the "curse of dimensionality," make real-time application difficult [18].

For traffic participants, particularly nearby vehicles, motion intention recognition has progressed through model-based techniques like Hidden Markov Models (HMM) [19] and data-driven methods, such as neural networks (e.g., LSTM, deep belief networks) [20], trained on datasets like NGSIM to better predict lane changes or lane-keeping actions.

To tackle the computational challenges of HJ analysis, our team suggests tapping into a large collision avoidance database to find approximate solutions, boosting efficient, safety-driven risk assessment while improving motion intention recognition for safer driving.

**LLMs-based Decision-making for autonomous Driving**: Recent work has highlighted how LLMs are making strides in autonomous driving decisions. Hu et al. [21] proposed an LLM-based misbehavior detection architecture that leverages fine-tuned large language models to identify forged traffic signs and motion data in autonomous vehicles. A VLM-MPC framework proposed by Long et al. [22] blends Vision Language Models with real-time control, using nuScenes data to keep vehicles safe in tricky settings like rain or busy intersections. Zheng el al.'s PlanAgent [23] uses a Multi-modal LLM to tackle rare, complex cases, beating top methods on nuPlan tests through step-by-step reasoning. Sha et al. [24] showed in LanguageMPC that LLMs can act as smart decision-makers, outperforming standard approaches in managing single and multiple vehicles with their practical know-how. Wang' et al. [25] explored LLMs for planning, adding a safety check to improve outcomes in unpredictable simulations, while Zhang et al. [26] proposed a ChatGPT-based framework that combines human-like reasoning and distributed control for robust collision avoidance in vehicle platoons. Huang et al. [27] presents SafeDrive, a risk-sensitive decision-making framework that integrates and data-driven approaches to enhance decision-making in high-risk and long-tail scenarios.

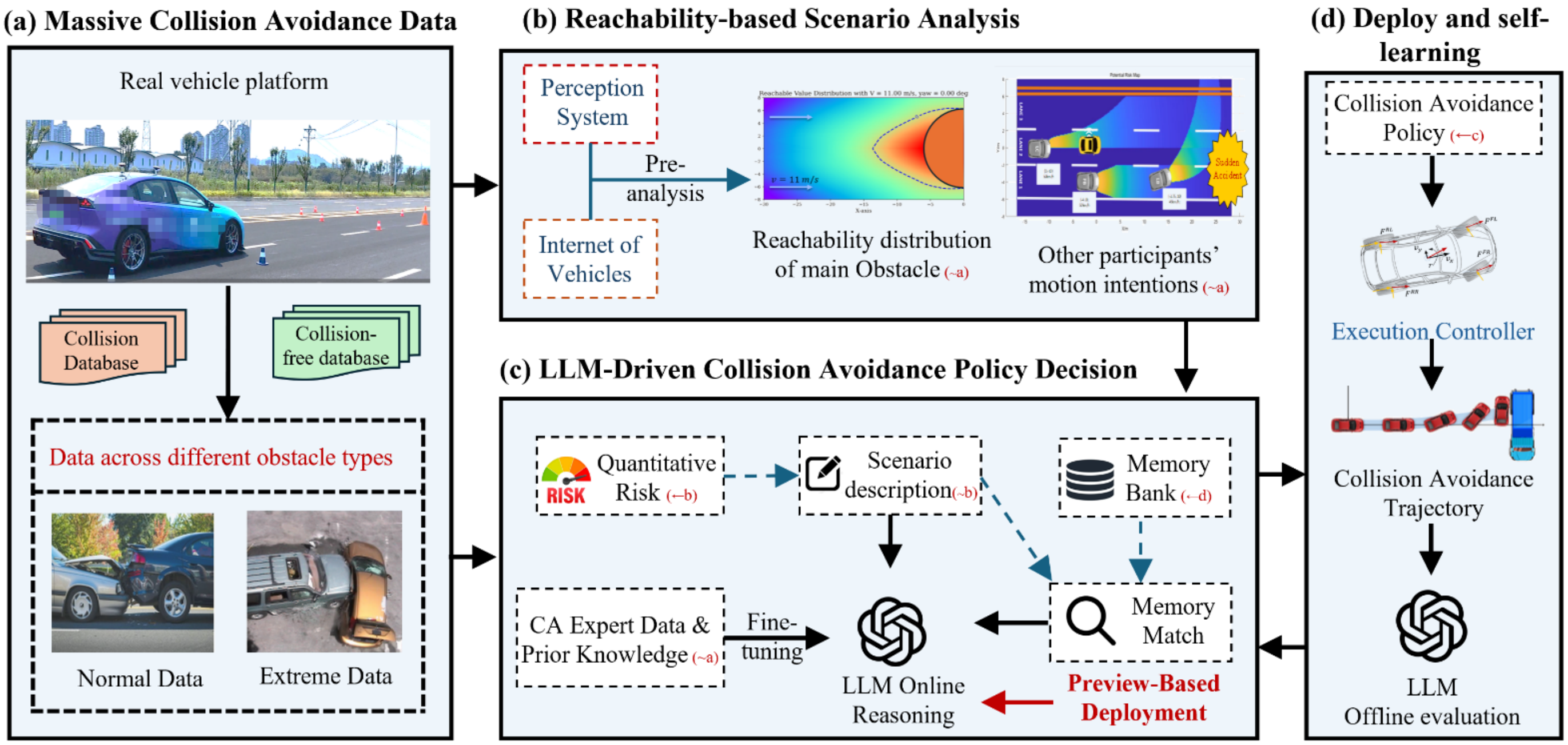

**Fig. 2.** Overview of the LLM-driven Scene-Aware Collision Avoidance (SACA) framework. (a) Real-vehicle collision avoidance dataset. (b) Reachability-based scenario analysis. (c) LLM-driven decision-making with risk assessment, expert knowledge, and memory retrieval. (d) Deployment and self-learning for continuous improvement. Arrows indicate direct outputs from a module, while the tilde (~) denotes related computations.

However, incorporating driving expertise and ensuring that LLMs operate smoothly in the real world remains a major hurdle, as some studies have pointed out that they require extensive traffic perception knowledge [28] and are slow to react at critical moments [29]. Our SACA framework steps in to address these gaps, using LLMs to achieve intelligent and practical collision avoidance in extreme driving situations.

## III. Framework

This study develops a Large Language Model (LLM)-driven Scene-Aware Collision Avoidance (SACA) upper-level decision-making framework, based on a vast dataset of real-vehicle collision avoidance scenarios and the team's pre-researched avoidance strategy library [4, 30-32], as illustrated in Fig. 2. The external interaction process of SACA involves receiving vehicle dynamics data, surrounding obstacle information, and neighboring traffic motion data. With the objective of minimizing collision loss, SACA integrates legal, ethical, and emotional considerations to determine the optimal collision avoidance strategy and its key parameters through scenario analysis, LLM online reasoning, and real-time deployment. The lower-level control module then executes the corresponding avoidance strategy, maneuvering the vehicle accordingly. The collision avoidance trajectory and impact outcomes are subsequently offline evaluated by the LLM and stored in a case library, facilitating self-learning within the SACA framework.

Table I presents the candidate collision avoidance strategies considered in this study. Policy 3 and Policy 4 represent coordinated braking and steering maneuvers, which do not involve high-slip conditions. These strategies primarily consist of lateral movement coupled with braking while ensuring rapid directional recovery. Additionally, SACA is expected to output the target velocity for these policies. Policy 5 and Policy 6, on the other hand, involve T-type drift maneuvers, where the vehicle deliberately induces controlled lateral sliding to rapidly adjust its orientation while decelerating. These strategies are particularly beneficial in scenarios with lateral oncoming traffic, as they help mitigate collision severity by redirecting the impact to the energy-absorbing front or rear zones, thereby protecting the occupant cabin and battery pack [33]. Regarding the internal workflow of SACA, the environment analysis module (Fig. 2b) performs reachability-based analysis of primary obstacles and predicts surrounding vehicle motion intentions, thereby constructing contextual risk awareness for decision-making.

TABLE I
Candidate Collision Avoidance Policies

| ID | Name | Description | Ref. |
|---|---|---|---|
| *0* | AEB | Engages maximum braking force to stop the vehicle. | [30] |
| *1* | AES-L | Executes a sudden left turn to change lanes and resume direction. | [31] |
| *2* | AES-R | Executes a sudden right turn to change lanes and resume direction. | [31] |
| *3* | ES-B-L | Changes lanes to the left while applying braking. | [4] |
| *4* | ES-B-R | Changes lanes to the right while applying braking. | [4] |
| *5* | T-D-L | Performs a T-type drift maneuver, bringing the vehicle to a perpendicular stop facing left. | [32] |
| *6* | T-D-R | Performs a T-type drift maneuver, bringing the vehicle to a perpendicular stop facing right. | [32] |
| *7* | NI | The situation does not require intervention. | - |

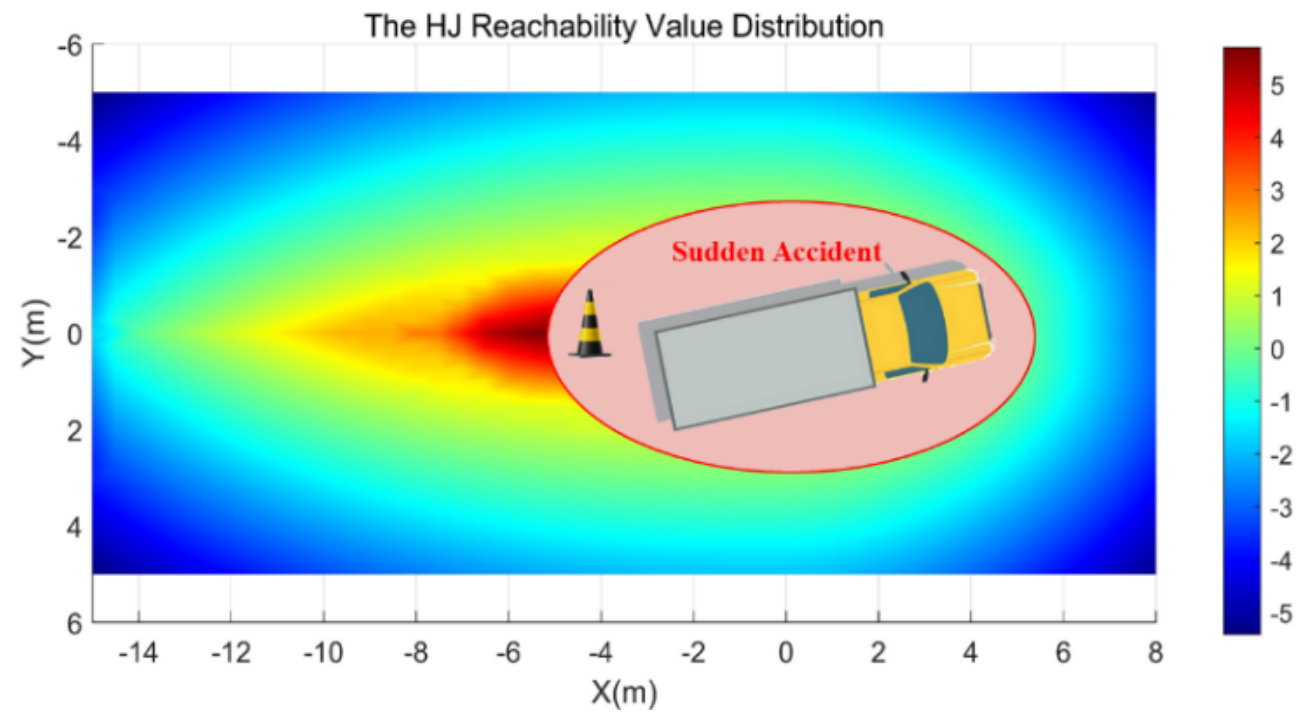

**Fig. 3**: HJ reachability value distribution for an elliptical obstacle. The red ellipse indicates the obstacle boundary. The color gradient represents reachability values, defined as Formula 1. Positive values indicate intrusion into the obstacle envelope, with higher values signifying greater risk.

Within the LLM-driven decision-making module (Fig. 2c), vector-based scene matching is conducted to compare the current scenario with stored cases in the memory bank, allowing the retrieval of historically effective collision avoidance results to serve as incremental prompts for decision-making. Additionally, expert-annotated avoidance examples and key safety constraints are leveraged to fine-tune the LLM, enhancing its adaptability to collision scenarios. The model then infers the optimal avoidance strategy based on the given context.

To address real-time constraints during deployment, a precomputed strategy approach is designed, enabling the previewing of scenarios and retrieving or reasoning strategies based on similarity and confidence levels, thereby improving the real-time performance of execution.

## III. METHODOLOGY

This section looks at how the SACA framework integrates specialized collision avoidance knowledge to support real-time, flexible decision-making. Essentially, it explains how to address the two major challenges of reliability and real-time deployment raised in Section I.

### *A. Predictive Scenario Analysis*

LLM requires input to describe scenarios incorporating collision avoidance expertise, thus requiring prior risk assessment and motion prediction to make informed decisions.

**Obstacle Reachability Analysis:** For obstacles with relatively fixed motion states, such as sudden accident vehicles, SACA uses Hamilton–Jacobi (HJ) reachability to gauge the danger they pose to the ego vehicle's operation. The reachable value, reflected in functions like $V_h(x)$ and $Q_h(x,u)$ indicates how well the vehicle can avoid such obstacles starting from the current state $x$, focusing solely on evasion—here, measured as a distance-based risk. $x$ represents the vehicle's kinematic and dynamic states while $u$ denotes vehicle controls (e.g., steering and throttle/braking) These functions are defined as:

$$V_h(x) = \min_{u(\cdot)\in\mathcal{U}}[\max_{\tau\in N} h(x_\tau)],\ x(0) = x \quad (1)$$

$$Q_h(x,u) = \min_{u(\cdot)\in\mathcal{U}}[\max_{\tau\in N} h(x_\tau)],\ x_0 = x, u_0 = u \quad (2)$$

where $h(x_\tau) \geq 0$ marks unsafe states (e.g., collisions), and $h(x_\tau) < 0$ denotes safety, with $x_\tau$ as the state at time $\tau$ under control $u \in U$. However, the high-dimensional computation of HJ reachability involving vehicle dynamics (known as the "curse of dimensionality") poses a challenge for real-time use [18]. To address this, we draw on a collision avoidance database—built from over 31 million real-world samples—and offline reinforcement learning [34,35] to approximate $V_h(x)$ and $Q_h(x,u)$ using neural networks, iteratively computing with a reachability Bellman operator:

$$\mathcal{B}^* Q_h(\boldsymbol{x},\boldsymbol{u}) \coloneqq (1-\gamma)h(\boldsymbol{x}) + \gamma\max\{h(\boldsymbol{x}), V_h^*(\boldsymbol{x}')\} \quad (3)$$

This approximate Bellman operator satisfies the contraction mapping property, enabling computation through value iteration. Offline training yields a neural network model for relevant obstacles, which, when deployed online, calculates their reachable values to assess risk levels in real time using $V_h(x)$. This process incorporates risk data accounting for the ego vehicle's operational capabilities into SACA framework. Figure 3, positioned here, illustrates the computed risk distributions for an elliptical obstacle. We normalize the computed reachable values to quantify risk levels across different obstacles, using:

$$R(x) = \frac{V_h(x) - V_h^{min}}{V_h^{max} - V_h^{min}}\ (mapped\ to\ [0,1]) \quad (4)$$

where $V_h^{max}$ and $V_h^{min}$ are manually set sensitivity adjustment parameters. This process incorporates risk data accounting for the ego vehicle's operational capabilities into the SACA.

**Traffic Participants Motion Intention Prediction**: Reachability analysis excels at assessing risks from obstacles with relatively stable motion states, leveraging the ego vehicle's dynamics and operational capabilities to gauge potential risk. However, for participants like surrounding vehicles, whose motion states can change rapidly, their intentions become an important factor. Consider two vehicles with identical positions and speeds: one intending to maintain its course poses minimal risk, while another planning an emergency brake could precipitate a collision. This difference highlights the necessity of incorporating motor intention into the input information of LLMs reasoning. We'll use surrounding vehicles as a prime example.
Collision risks often stem from sudden longitudinal or lateral maneuvers by surrounding vehicles, such as lane changes or braking. To address this, we propose an intention recognition module that classifies the motion intentions of surrounding vehicles (I) into six categories: Maintain (M), forward braking (FB), left braking (LB), left lane change (LC), right braking ($RB$), and right lane change ($RC$). This is formalized as:

$$\mathrm{I} \in \mathrm{I_{all}} = [\mathrm{M, LC, RC, FB, LB, RB}]$$

where $I_{all}$ denotes the set of all possible intentions.

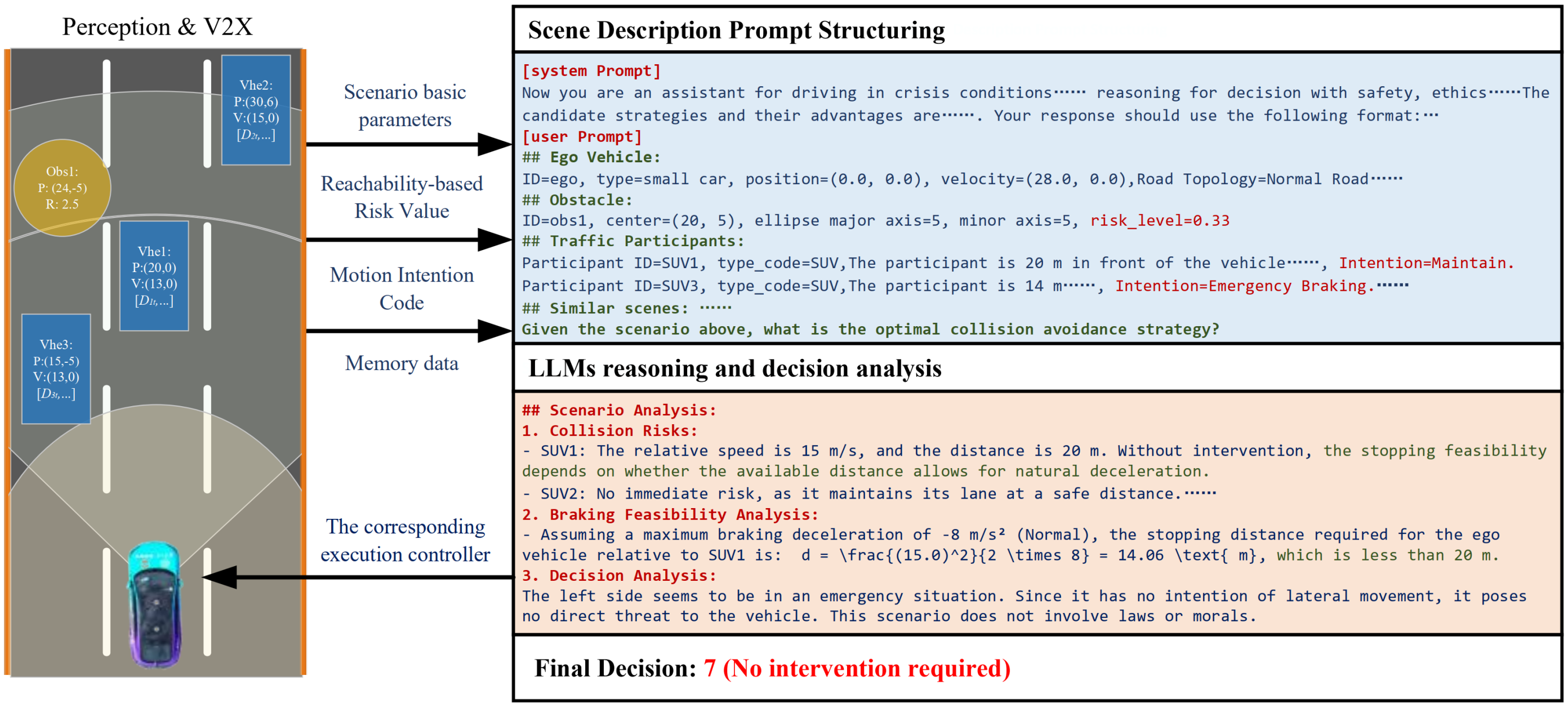

**Fig. 4.** An example illustrating how essential information in a driving scenario—such as the vehicle's position, surrounding risks, and the intentions of nearby traffic—is organized in the top section. The middle section walks through the thinking behind using that information to suggest what to do next. The bottom part wraps it up with the final decision.

We predict these intentions using a hybrid Long Short-Term Memory (LSTM) and Conditional Random Field (CRF) model, leveraging recent time-series driving behaviors like speed, steering, and pedal inputs over recent moments. The LSTM processes this sequence to estimate intentions, expressed as:

$$\widetilde{I_t} = LSTM(D_t, D_{t-1}, \dots)$$

where $D_t$ represents the driving input data at time $t$. However, LSTM alone overlooks logical transitions between consecutive intentions—e.g., a vehicle is unlikely to shift from left braking (LB) to right braking (RB) abruptly. The CRF addresses this by enforcing consistency through a transition probability matrix $A$, where $a_{ij} = P(I_t = I_{\mathrm{all}}\,(j) \mid I_{t-1} = I_{\mathrm{all}}\,(i))$, derived from expert knowledge [4]. The final output, $I_t$ retains LSTM's sensitivity to continuous driving patterns while incorporating real-world constraints on adjacent intention transitions.

The LSTM-CRF process supports detection of sudden lane changes or braking, enhancing collision risk assessment when combined with reachability analysis. While this study focuses on vehicles, the approach can be extended to other road users with appropriate motion models.

**Prompt Organization for Scenario Description:** The SACA framework uses a structured scene description prompt to guide LLMs in making effective collision avoidance decisions. This prompt integrates the ego vehicle's state (e.g., position [0, 0], velocity 28 m/s), obstacle risk profiles (e.g., small car, reachability-based risk 0.33), and traffic participant intentions (e.g., Maintain or Emergency Braking) from HJ reachability and LSTM-CRF analyses.

As shown in the upper section of Figure 4, the prompt details the ego vehicle's context, obstacle metrics (e.g., 5 m axis), and participant actions (e.g., emergency braking), facilitating memory bank retrieval or LLMs reasoning.

*B. Fine-Tuning and Case-Based Reasoning*

We introduce a dual strategy that combines supervised fine-tuning (SFT) with case-based reasoning (CBR) into SACA.

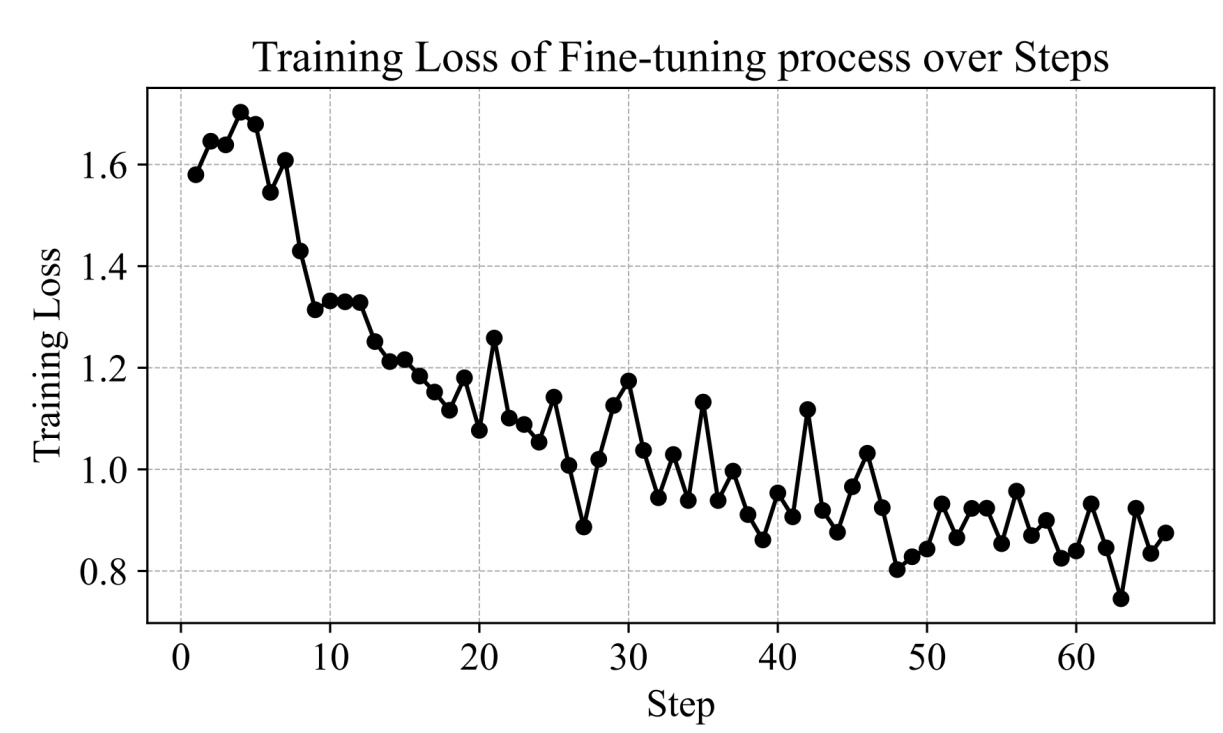

**Fig. 5.** Training loss curve for fine-tuning *GPT-4o-mini-2024-07-18*, illustrating consistent convergence over the training steps.

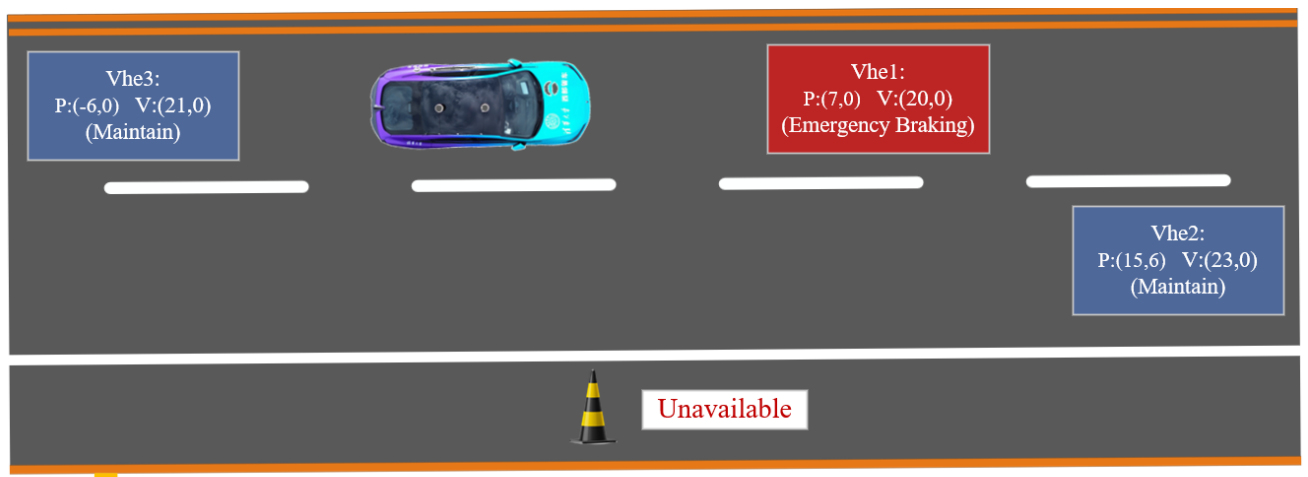

Using the same scenario description (User Input).

| Model | Decision | Speed Target | Reasoning |
|---|---|---|---|
| Grok | Left lane change with braking (Action 3) | 15 m/s | Mistakenly considered the right side unavailable |
| OpenAI baseline | T-type drift avoidance (Action 5) | Not specified | Excessive and incorrect reliance on experiential input in cue words |
| **Fine-tuned model** | Right lane change with braking (Action 4) | 10 m/s | Safe clearance considering relative positions and speeds |

**Fig. 6.** Decision outcomes of Grok, OpenAI baseline, and the fine-tuned LLM in a frontal emergency-braking scenario.

through fine-tuning, the LLMs in SACA internalize domain-specific collision avoidance knowledge, physics-based rules, and common scenario data, enabling it to function like a seasoned "collision avoidance expert." Complementing this, the CBR serves as a "living notebook," allowing retrieval of past successes and failures to guide decisions in rare or extreme conditions.

We performed supervised fine-tuning on the *GPT-4o-mini-2024-07-18* model using OpenAI's recommended pipeline and hyperparameter settings. The training corpus consisted of domain-specific question–answer pairs derived from over 50 collision cases selected with reference to the NHTSA Fatality Analysis Reporting System (FARS), covering various highway, intersection, and ramp scenarios, along with expert reasoning demonstrations (Will be made public in the project files after publishing). In total, we used 235,284 training tokens, which met the fine-tuning guidelines of OpenAI. Figure 5 illustrates the training loss curve over successive steps, showing a consistent decline and effective convergence.

Compared to baseline *GPT-4o-mini* and *Grok* models, our fine-tuned LLMs exhibits notably improved spatial perception and domain-specific collision avoidance knowledge, such as estimating computing break distances in Figure 4's reasoning process. Figure 6 demonstrates that all three models were given identical scenario descriptions and system prompts involving expert knowledge. Despite this, the Grok model selects policy 3 (a left lane-change with braking), and the OpenAI baseline proposes policy 5 (a T-type drift maneuver). These choices indicate a misunderstanding of spatial constraints and an inappropriate use of expert guidance. Our model with fine-tuning correctly evaluates the spatial limitations and the risk presented by the trailing vehicle, choosing policy 4 (a right lane-change with braking) to preserve a safe distance.

To enhance the proposed method, we incorporated a case-based reasoning mechanism, as depicted in Figure 2. Each time SACA is activated, an independent LLM retrospectively assesses the collision avoidance performance and evaluates collision losses. This process enables continuous improvement through online case-based self-learning.

### *C. Scenario-Preview-Based Deployment*

We developed a scenario-preview-based deployment, as illustrated in Fig. 7. This method initiates when the calculated time-to-collision (TTC) drops below a predefined threshold ($TTC < T_2$), prompting scenario previewing based on the current obstacle dynamics and predicted motion intentions of surrounding traffic participants.

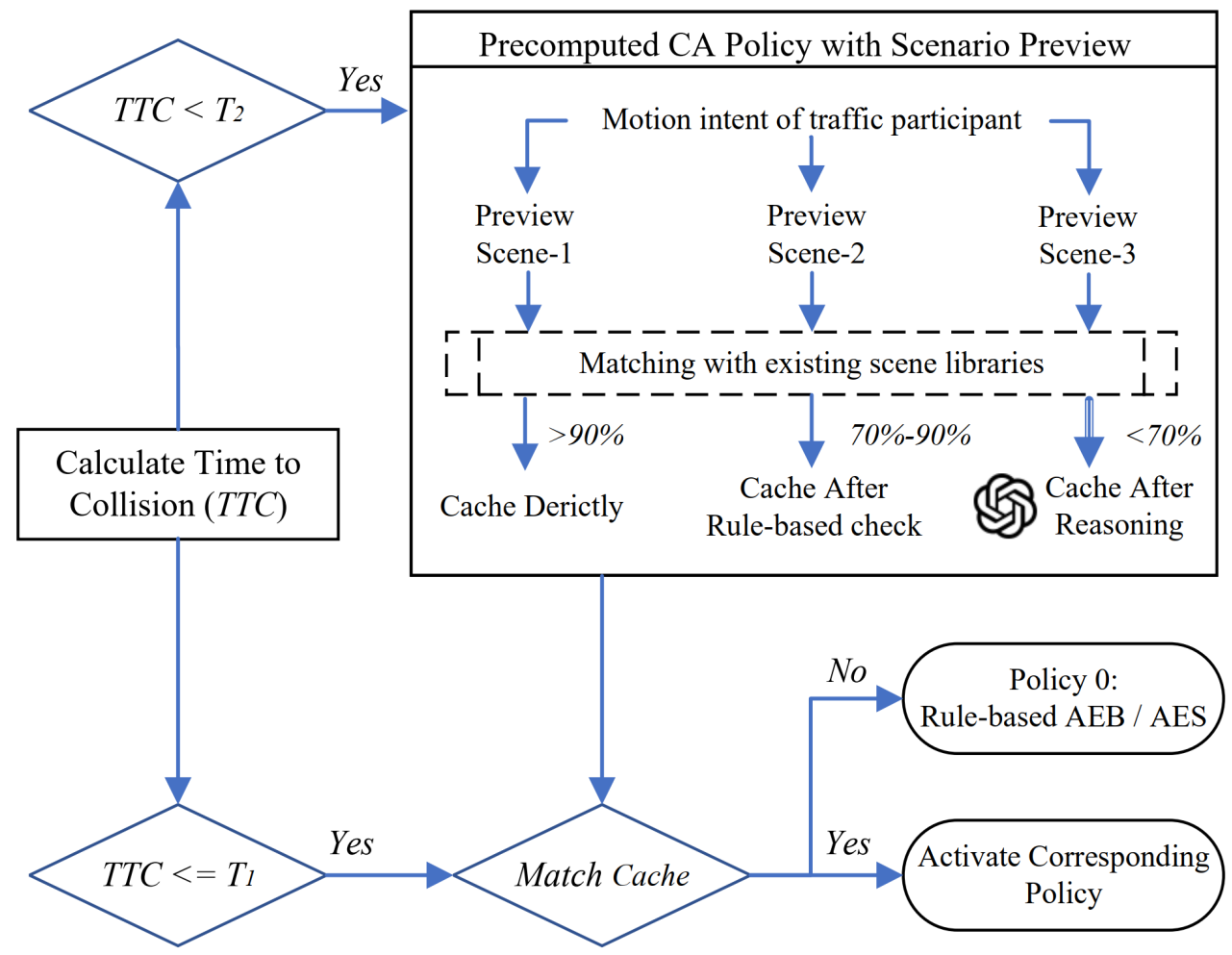

**Fig. 7.** Scenario-preview-based deployment mechanism in SACA.

---

**Algorithm 1:** Scenario-Aware Collision Avoidance Integrating LLMs-Driven Reasoning

---

**Input:** Road geometry data, obstacle info (position & dimensions), surrounding vehicles' states (position, velocity, controls), thresholds $T_1, T_2$, memory bank, fine-tuned LLM
**Output:** Selected collision avoidance **policy ID**, updated memory bank after action and evaluation
**Compute** current Time to Collision (TTC);
**if** $TTC < T_2$ **then**
  **Conduct** obstacle reachability analysis (e.g., Hamilton–Jacobi);
  **Predict** motion intentions of surrounding vehicles (e.g., lane change, braking);
  **Forecast** three potential collision scenarios at future time $T_1$;
  **Generate** structured scenario descriptions for each forecasted scenario;
  **foreach** *forecasted scenario* **do**
    **Compare** with existing scenarios in memory bank to get a similarity score;
    **if** *similarity > 90%* **then**
      **Cache** the corresponding **policy** immediately
    **else if** *70% < similarity ≤ 90%* **then**
      **Perform** safety validation check, then **cache** the **policy**
    **else**
      **Use** fine-tuned LLM reasoning with the nearest scenario from memory bank;
      **Cache** the **policy** inferred by the LLM
    **end**
  **end**
  **if** *current* $TTC \leq T_1$ *(collision avoidance trigger)* **then**
    **Match** the current scenario with cached scenarios;
    **if** *match similarity > 90%* **then**
      **Execute** the cached collision avoidance **policy**
    **else**
      **Execute** the default rule-based emergency brake (**Policy 0**: AEB/AES)
    **end**
  **end**
**end**
**After execution:** perform offline evaluation of the trajectory and any collision outcomes using a dedicated LLM;
**Update** the memory bank with new scenario data and evaluation results for continuous improvement;

---

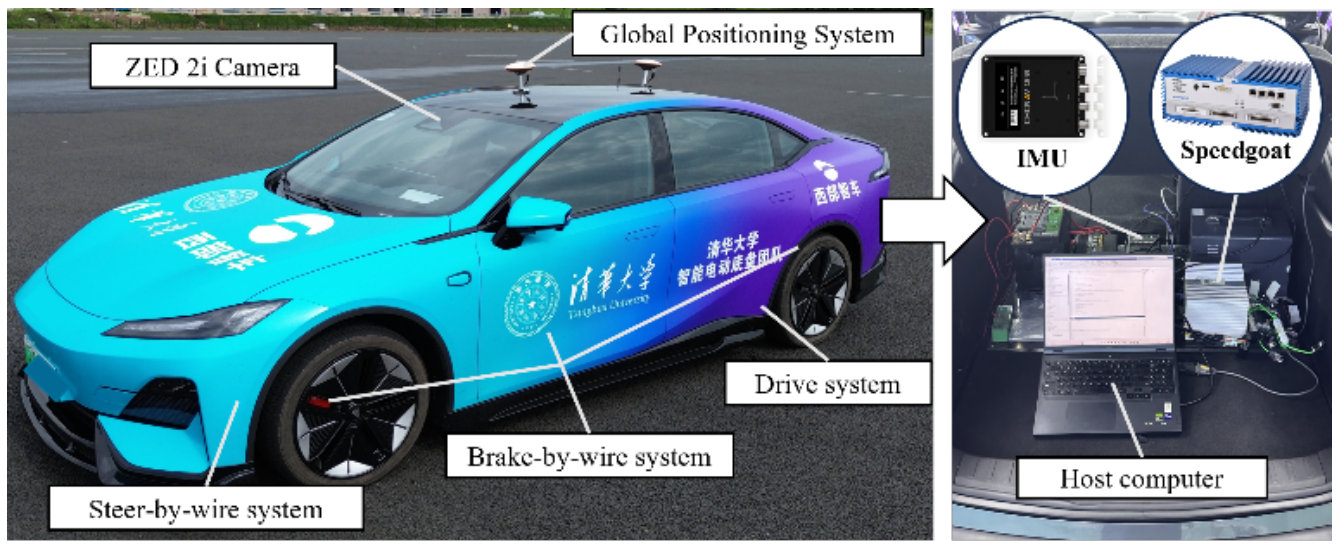

**Fig. 8.** Test vehicle platform and its key components

TABLE II THE EMPLOYED VEHICLE PARAMETERS

| Parameter | Value | Parameter | Value |
|---|---|---|---|
| Total Mass ($kg$) | 1615 | Initial Speed ($m/s$) | 13 |
| Roll inertia ($kg.m^2$) | 251 | Friction coefficient | 0.75-0.9 |
| Yaw inertia ($kg.m^2$) | 2795 | Tire | 215/55-R17 |

Specifically, the framework anticipates three likely collision scenarios at future time instant T_1, which are simultaneously matched against scenarios stored in the memory bank. If a predicted scenario achieves a similarity above 90% with an existing scenario, the corresponding collision avoidance strategy is directly cached without additional evaluation. When the similarity lies between 70% and 90%, a rule-based safety check is performed to validate the strategy before caching. If similarity is below 70%, LLM-driven online reasoning is invoked to infer an appropriate collision avoidance strategy, which is then cached.

At the actual collision avoidance decision time ($TTC < T_1$), the real-time scenario is compared against the cached previews. If a matching cached scenario is identified, its associated collision avoidance policy is immediately activated. In the absence of a match, the system defaults to a rule-based automatic emergency braking (AEB/AES) strategy.

This preview-based deployment improves the real-time applicability of SACA by allowing more inference time (the interval between $T_2$ and $T_1$) for LLMs, as confirmed through tests with OpenAI API in the fastest response case. It is important to note that, although this method substantially enhances deployability and shows real-time responsiveness under optimal API conditions ( $T_2 - T_1 > 3s$ ), current computational resources still limit consistently reliable LLM deployment. Nevertheless, compared to former studies [27-29], this framework provides a feasible pathway for integrating LLM-driven decision-making into critical, time-sensitive collision avoidance scenarios.

In summary, SACA combines scenario prediction, logical reasoning, and memory-based retrieval, with its execution process outlined in the pseudocode above.

## IV. EXPERIMENTAL EVALUATION

To validate the effectiveness and deployability of the proposed SACA framework, real-world tests were conducted in critical collision scenarios.

**Preview Scenario at $T_1$**

```
AUTONOMOUS DRIVING COLLISION AVOIDANCE SCENARIO
## Ego Vehicle:
ID=ego, type=small car, position=(0.0, 0.0), velocity=(14.0,
0.0),Road Topology=Intersection

## Obstacles:
No static obstacles detected.

## Traffic Participants:
Participant ID=object1, type=Large Truck.
The participant is 16.0 m in front of the vehicle and 6.4 m
to the left.
The participant is moving to the right at a speed of 7.5 m/s.
Intention = FD (Maintain).

Participant ID=pedestrian, type=pedestrian
The participant is 6 m in front of the vehicle and 4 m to the
left.
The participant is moving to the left at a speed of 0 m/s.
Intention = Maintain.

## HISTORICAL SCENARIOS:
No historical data is available.
```

**SACA Reasoning Process**

```
Scenario Analysis
1. Ego Vehicle State:
- The ego vehicle is traveling at 14 m/s in an intersection.

2. Collision Target (Large Truck):
- Located 16.0 m ahead and 6.4 m to the left.
- Moving right at 8.0 m/s, crossing the ego vehicle's path.
- TTC Calculation: TTC = 16/14 = 1.23s
- Since TTC = 1.3s, intervention is required.

4. Collision Risk Assessment:
- The truck's lateral movement means it will cross the ego
vehicle's path imminently.
- The pedestrian's position limits the use of leftward
avoidance maneuvers.

5. Decision Analysis:
- Braking alone (Action 0) may not prevent a
collision, ……..
- Leftward maneuvers (Actions 1, 3, 5) are not viable due to
the pedestrian.
- Rightward maneuvers (Actions 2, 4) are not viable, as they
may not provide enough time to avoid …….
- The impact should be directed to the energy-absorbing
structure to protect the battery and occupants.
- T-type drift avoidance to the right side (Action 6) is the
safest option.

Response to user: 6
```

**Fig. 9.** Validation of the SACA reasoning process in the intersection scenario.

### A. Experimental Setup

The verification platform is a full-scale consumer-grade rear-wheel drive electric vehicle, shown in Figure 8. It features a steer-by-wire system for closed-loop front wheel control, a drive system for torque control, and a brake-by-wire system for master cylinder pressure. Key specs like mass and tire details are in Table III, courtesy of the manufacturer. Cameras, inertial navigation (CGI 610), and GNSS track position, velocities, accelerations, yaw rate, and wheel speed in real time. Data flows via CAN FD to a real-time computer (*Speedgoat*), which processes it and sends control commands, powered by an Intel i7 2.5 GHz dual-core CPU, 8 GB RAM, and high-speed I/O613 modules.

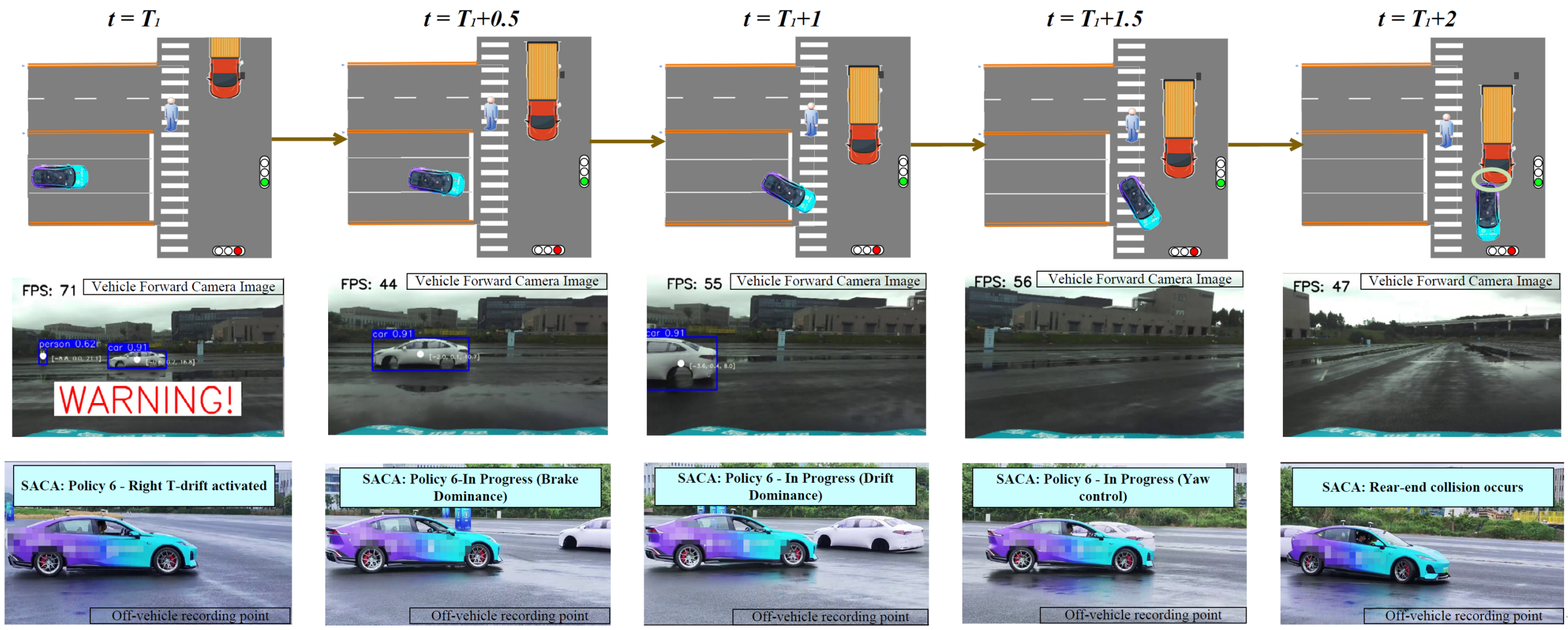

**Fig. 10** Real-vehicle experimental validation of the T-type drift collision avoidance maneuver in the intersection scenario.

Because the main novelty of this work is SACA's scenario-based arbitration for extreme collision avoidance, we compared its performance to two baseline decision-making approaches:

- **LLMs without Fine-tuning:** retains the same overall system architecture but omits domain-specific fine-tuning of LLMs, relying only on the general capabilities.
- **Imitation Learning:** is an approach inspired by [36] that directly trains a policy on expert demonstration data for high-risk collision scenarios, without incorporating scenario-based reasoning or preview mechanisms.

For overall test safety, the collision avoidance strategy initiation threshold (TTC trigger time $T_1$) was set to 1.3 seconds. Considering the response latency of the fine-tuned LLM model, scenario previewing and inference initiation threshold (TTC trigger time $T_2$) was set at 5.5 seconds. These parameters ensure reliable evaluation of SACA's real-time decision-making capabilities within current OpenAI API response constraints.

*B. Intersection Scenario*

In the intersection scenario, a mixed virtual-real approach was adopted for safety reasons. Specifically, an actual stationary vehicle positioned transversely simulated a high-speed lateral incoming vehicle. Upon detection by the ego vehicle's perception system, this stationary vehicle was treated as dynamically approaching from the lateral direction at high speed.

At the time instant $T_2 = 5.5s$ triggering scenario preview, the ego vehicle is traveling toward an intersection at $14\,m/s$. A forward-facing ZED 2i camera detects an inflatable car at approximately $55\,m$ ahead, and the virtual environment indicates a truck located $55\,m$ ahead and $43\,m$ on the left, moving laterally at $7.5\,m/s$ with a "maintain" motion intention. Additionally, a pedestrian is detected at roughly $48\,m$ forward and $8.2\,m$ to the left, crossing from left to right at $1\,m/s$.

Figure 9 presents a unified view of the scenario preview result and the corresponding LLM-driven reasoning workflow in SACA. In the upper portion, the ego vehicle's environment at time ($T_2$) is outlined, including the positions, speeds, and predicted intentions of surrounding objects and pedestrians. In the lower portion, the diagram shows how the fine-tuned LLM processes the scenario prompt to get a collision avoidance decision. The preview reveals the large truck will cross the ego vehicle's path in approximately 1.3 s. The pedestrian's proximity rules out leftward maneuvers, and simple braking does not provide adequate deceleration to avoid collision. Consequently, SACA recommends a T-type drift to the right (Action 6), which reorients the vehicle to mitigate impact by directing collision forces into the energy-absorbing structures and preserving occupant safety.

Figure 10 presents the real-vehicle execution results of the T-type drift collision avoidance maneuver under the intersection scenario. The schematic (top row) illustrates how the relative positions of the truck, pedestrian, and ego vehicle evolve from $t = T_1$ to $t = T_1 + 2$. Forward-camera images (middle row) confirm real-time detection of the truck and pedestrian, triggering an explicit warning before initiating the avoidance action. Off-vehicle recordings (bottom row) validate the lateral displacement and controlled drift, clearly demonstrating how the ego vehicle rapidly adjusts its orientation to redirect collision forces into the rear energy-absorbing structures, thereby effectively avoiding injuries to occupants and pedestrians.

We conducted comparative tests and ablation study using the two baseline methods without relying on any pre-cached scenarios. The ego vehicle's initial speed was randomly varied between 10–15 m/s, and the truck was initialized with two motion states: forward driving (FD) representing a high-risk scenario, and emergency braking (EB) representing a no-collision-risk scenario. Each test case was repeated ten times to ensure robustness.

Two performance metrics were tracked:

- **Average Collision Loss**: Estimated based on the relative

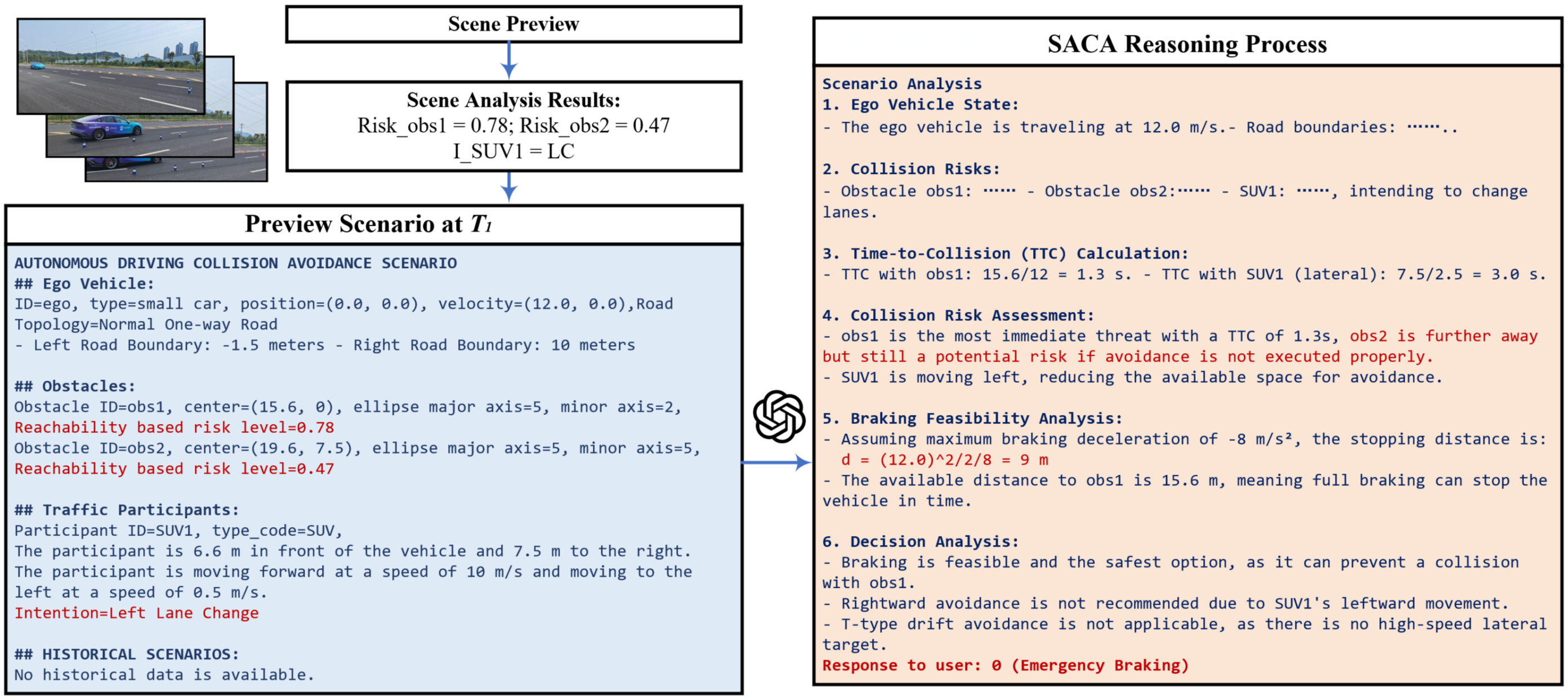

**Fig. 11.** Scene preview and LLM-driven reasoning workflow under a one-way road scenario.

TABLE III ABLATION COMPARISON OF COLLISION LOSS AND FALSE-TRIGGER LOSS ACROSS TESTED METHODS.

| Methods | Collision Loss | False-Trigger Loss |
|---|---|---|
| **Imitation Learning [38]** | 2.36 | 0.35 |
| **LLMs without Fine-tuning** | 1.85 | 0.27 |
| **LLMs without Scenario-Awareness** | 0.61 | 0.39 |
| **Our SACA** | **0.54** | **0.05** |

velocity, reflecting impact severity. To account for varying injury risks, location-specific injury rates are applied [37]: 0.41 for side collisions, 0.35 for front-end collisions, and 0.21 for rear-end collisions. The final loss is obtained by scaling the collision energy with the corresponding injury risk factor.

➢ **False-Trigger Loss:** In no - collision - risk scenarios, mistakenly triggering moderate maneuvers (Policy 0–4) incurs 0.5 cost; incorrectly activating extreme drift maneuvers (Policy 5–6) incurs 1.0 cost. The average cost across trials represents the overall false-trigger rate.

The compared methods include: (1) Imitation Learning, a behavior cloning baseline trained from expert demonstrations without involving LLMs; (2) LLM without Fine-tuning, which uses a general-purpose LLM without domain-specific adaptation; and (3) LLM without Scenario-Awareness, which removes the risk assessment and motion intention prediction modules from the input prompt, leaving only raw kinematic states. The evaluation results are summarized as Table III.

Table III confirms the benefit of both fine-tuning and scene-aware perception. Our full SACA pipeline records the lowest collision loss (0.54) and almost suppresses false triggers (0.05). Removing fine-tuning (LLM w/o FT) raises collision loss to 1.85 and leaves a non-negligible false-trigger loss of 0.27, showing that adaptation to driving data is essential. Discarding scene-aware cues while keeping the fine-tuned model (LLM w/o SA) further lowers collision loss to 0.61 but at the expense of the highest false-trigger loss (0.39), indicating unstable decisions when contextual risk information is missing. The traditional imitation-learning baseline performs worst overall, with the largest collision loss (2.36) and moderate false-trigger incidents (0.35). These results demonstrate that fine-tuning and structured scene awareness act synergistically: both are required to achieve reliable and precise collision-avoidance decisions in extreme conditions.

*C. One-way Road Scenario*

In this scenario, when the time-to-collision $T_2 = 5.5\,s$ is reached, the ego vehicle travels along a one-way road at $12\ m/s$ . An elliptical obstacle with a major radius of $3.5\,m$ and a minor radius of $1.75\,m$ is located $66\,m$ ahead in the same lane. To the right, at an offset of $7.5\,m$ and $70\,m$ forward, is a circular obstacle with a radius of $5\,m$. The road's left boundary is $1.5\ m$ from the ego lane, while the right boundary sits $10\ m$ away, forming two additional lanes. A surrounding vehicle—positioned $7.5\ m$ to the right and $15\ m$ forward—is moving at $10\,m/s$ with an intention to merge left.

In our real-vehicle tests, red barriers were placed to simulate obstacles and blue barriers to represent lane boundaries and the trajectory of vehicles changing lanes. The subsequent sections compare how baseline methods and our SACA framework respond to potential obstacles and lane changes in the one-way road environment.

Figure 11 provides an overview of the scene preview, its corresponding risk assessment, and the LLM-driven reasoning process. From the top-level analysis, obstacle obs1 (TTC ≈ 1.3 s) poses the most immediate collision threat, while obstacle obs2 is less urgent but still requires caution if not addressed properly. The SUV, labeled as "I_SUV1," is shifting left at a moderate speed, further constraining the available space. Based on reachability-based risk values (0.78 for obs1 and 0.47 for obs2) and a braking feasibility check, SACA's decision module determines that maximum braking (Policy 0) is sufficient to stop the vehicle before reaching obs1, thus mitigating a potential collision.

TABLE IV COMPARISON OF COLLISION LOSS AND FALSE-TRIGGER LOSS ACROSS TESTED METHODS.

| Methods | Collision Loss | False-Trigger Loss |
|---|---|---|
| **Imitation Learning [38]** | 0.49 | 0.50 |
| **LLMs without Fine-tuning** | 1.06 | 0.20 |
| **LLMs without Scenario-Awareness** | 0.65 | 0 |
| **Our SACA** | **0.07** | **0** |

We conducted comparative tests using the baseline methods described in Section IV.A under the one-way road scenario with identical preview settings. The ego vehicle's initial velocity was fixed at 12 m/s, and the positions of the obstacles and surrounding vehicles were initialized as previously described. Each method underwent 10 repeated trials in collision-risk and no-collision-risk configurations. The results are summarized in Table IV.

Quantitative analysis shows that our SACA achieves the lowest average collision loss (0.0697), significantly lower than all baseline methods. The imitation learning approach triggers identical maneuvers in all trials (Policy 4), consistently resulting in minor collisions at low speed (1.2 m/s) and thus leading to a higher collision loss (0.492). The LLM without fine-tuning exhibits diverse maneuver choices, leading to inconsistent outcomes and the highest collision loss (1.06). Although LLM without scenario-awareness achieves moderate collision loss (0.656), it frequently selects maneuvers lacking integrated braking, reducing effectiveness in collision avoidance.

In false-trigger tests, SACA and LLMs without scenario-awareness correctly identify scenarios requiring no intervention, resulting in zero false-trigger loss. However, imitation learning consistently misinterprets scenarios, repeatedly performing unnecessary lane-change maneuvers (Policy 4) and accruing a moderate false-trigger loss (0.50). LLM without fine-tuning occasionally misjudges scenarios, causing unnecessary interventions (Policies 2 and 4) and incurring a lower yet non-zero false-trigger loss (0.20).

### *D. Real-Time Feasibility*

In this set of experiments, the predicted collision times $T_2 = 5.5\,s$ (start of scenario preview) and $T_1 = 1.3\,s$ (initiation of collision avoidance) imply a maximum allowable LLM reasoning duration of $4.2\,s$. As discussed in Section III.C, current computational constraints can still pose reliability challenges for LLM-based approaches. However, under favorable OpenAI API response conditions, the proposed system can generate an avoidance policy within the allocated $4.2\,s$.

Inference times and token lengths are summarized in Table

TABLE V COMPARISON OF COLLISION LOSS AND FALSE-TRIGGER LOSS ACROSS TESTED METHODS IN ONE-WAY ROAD SCENARIOS.

| Scenario | Avg. Response Token | Avg. Response Time (s) | Min. Response Time (s) |
|---|---|---|---|
| Intersection Scenarios | 371 | 5.7 | 3.7 |
| One-Way Road Scenario | 352 | 5.3 | 3.5 |

IV. Only episodes completing inference within 4.2 s were considered for previous performance analyses. Although average inference times exceed 4.2 s, the minimum times confirm that SACA can reliably meet real-time constraints under optimal API response conditions.

.

## V. DISCUSSIONS

This study proposes the Scenario-Aware Collision Avoidance (SACA) framework, integrating Large Language Models (LLMs), predictive risk assessment, and memory-based decision-making for autonomous vehicles. SACA uniquely integrates ethical and legal factors with collision avoidance expertise, achieving a 70.8%–93.4% reduction in collision loss and an 81.5%–100% reduction in false-trigger loss compared to baselines in tested scenarios, while maintaining real-time performance under optimal API response conditions. The primary limitation remains the inference speed of LLMs. Future research should explore edge computing architectures or regional cloud computing centers to optimize model efficiency, enhance scenario databases, and integrate advanced perception systems, further improving performance and applicability. Overall, integrating LLMs into intelligent transportation systems holds great potential, promising safer and more ethically aligned autonomous driving solutions.

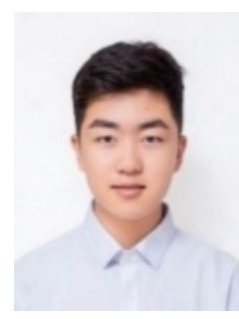

**Shiyue Zhao** is currently a visiting scholar at the University of Michigan, Ann Arbor, USA. He received the B.S. degree in traffic engineering from Central South University, Hunan, China, in 2021, and is pursuing the Ph.D. degree at the School of Vehicle and Mobility, Tsinghua University, Beijing, China. His research interests include transportation systems, advanced vehicle control, and vehicle extreme control.

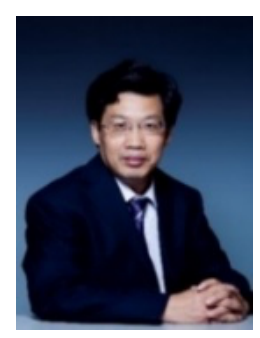

**Junzhi Zhang** received the B.S., M.S., and Ph.D. degrees in transportation from Jilin University, Changchun, China, in 1992, 1995, and 1997, respectively. He is currently a Professor in Power Engineering and Engineering Thermophysics with the State Key Laboratory of Automotive Safety and Energy, Tsinghua University, Beijing, China.

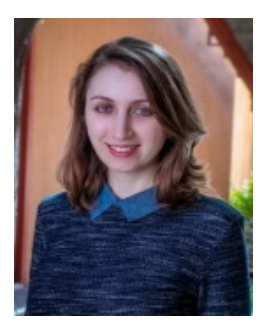

**Neda Masoud** received her B.S. in Industrial Engineering from Sharif University of Technology, M.S. in Physics from the University of Massachusetts, and Ph.D. in Civil Engineering from the University of California, Irvine. She is currently an Associate Professor in Civil and Environmental Engineering at the University of Michigan, Ann Arbor.

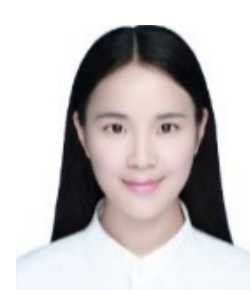

**Heye Huang** received the B.E. degree from Central South University, Changsha, in 2018, and the Ph.D. degree from Tsinghua University in 2023. She is currently a Research Associate Fellow with Connected & Autonomous Transportation Systems Lab, University of Wisconsin–Madison. Her current research interests include connected and automated vehicles, decision making and motion prediction.

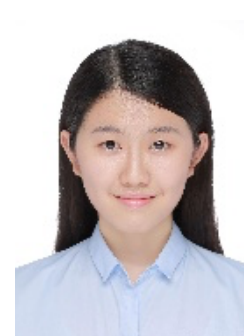

**Xiaohui Hou** received the B.S. and PH.D. degrees in School of Vehicle and Mobility from Tsinghua University in 2017 and 2022. She is now a postdoctoral fellow at the School of Automation at the Beijing Institute of Technology. Her research interests include autonomous vehicle decision making and motion planning.

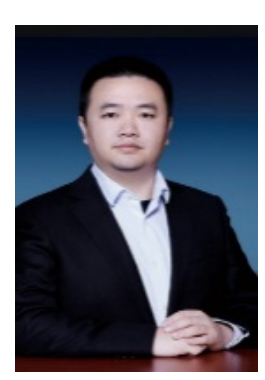

**Chengkun He** received the B.S. degree in Mechanical Design and Automation from Tsinghua University in 2008 and the Ph. D degree from Institute of Electrical Engineering Chinese Academy of Sciences, Beijing, China, in 2014. He is currently a Research Assistant in the State Key Laboratory of Automotive Safety and Energy, Tsinghua University, Beijing, China. His research interests include autonomous vehicle test, and Intelligent car chassis